\documentclass[runningheads,a4paper]{llncs}
\usepackage[pdftex]{graphicx}
\usepackage{amsmath,latexsym,amssymb}
\usepackage{hyperref}
\usepackage{booktabs}
\usepackage{caption}
\usepackage{tabularx}
\usepackage[usenames]{color}
\usepackage{indentfirst}
\usepackage{pbox}
\usepackage{caption}
\newcommand{\zm}[1]{\textcolor{black}{#1}}

\DeclareMathOperator*{\argmin}{arg\,min}

\def\etal{et al.}

\newcommand{\fig}[1]{Fig.~#1}
\newcommand{\tab}[1]{Table~#1}

\newcommand{\cmmnt}[1]{}

\begin{document}
\title{Learning to Segment Skin Lesions\\ from Noisy Annotations}
\titlerunning{}
\authorrunning{Mirikharaji, Yan and Hamarneh.}
\author{Zahra Mirikharaji, Yiqi Yan, and Ghassan Hamarneh}

% index{Mirikharaji, Zahra}
% index{Hamarneh, Ghassan}
\institute{School of Computing Science, Simon Fraser University, Canada\\
\email{\{zmirikha, yiqiy, hamarneh\}@sfu.ca}\\}

\maketitle

\begin{abstract}
Deep convolutional neural networks have driven substantial advancements in the automatic understanding of images. Requiring a large collection of images and their associated annotations is one of the main bottlenecks limiting the adoption of deep networks. In the task of medical image segmentation, requiring pixel-level semantic annotations performed by human experts exacerbate this difficulty. This paper proposes a new framework to train a fully convolutional segmentation network from a large set of cheap unreliable annotations and a small set of expert-level clean annotations. We propose a spatially adaptive reweighting approach to treat clean and noisy pixel-level annotations commensurately in the loss function. We deploy a meta-learning approach to assign higher importance to pixels whose loss gradient direction is closer to those of clean data. Our experiments on training the network using segmentation ground truth corrupted with different levels of annotation noise show how spatial reweighting improves the robustness of deep networks to noisy annotations.

\end{abstract}

%\begin{keywords}
%\end{keywords}
%

\section{Introduction}

Skin cancer is one of the most common type of cancers, and early diagnosis is critical for effective treatment~\cite{siegel2017cancer}. In recent years, computer aided diagnosis based on dermoscopy images has been widely researched to complement human assessment. Skin lesion segmentation is the task of separating lesion pixels from background. Segmentation is a nontrivial task due to the significant variance in shape, color, texture, etc. Nevertheless, segmentation remains a common precursor step for automatic diagnosis as it ensures subsequent analysis (i.e. classification) concentrates on the skin lesion itself and discards irrelevant regions.

Since the emergence of fully convolutional networks (FCN) for semantic image segmentation~\cite{long2015fully}, FCN-based methods have been increasingly popular in medical image segmentation. Particularly, U-Net~\cite{ronneberger2015u} leveraged the encoder-decoder architecture and applied skip-connections to merge low-level and high-level convolutional features, so that more refined details can be preserved. FCN and U-Net have become the most common baseline models, on which many different proposed variants for skin lesion segmentation were based. Venkatesh \etal~\cite{venkatesh2018deep} and Ibtehaz \etal~\cite{ibtehaz2019multiresunet} modified U-Net, designing more complex residual connections within each block of the encoders and the decoders. Yuan \etal~\cite{yuan2017automatic} and Mirikharaji \etal~\cite{mirikharaji2018star} introduced, in order, a Jaccard distance based and star-shape loss functions to refine the segmentation results of the baseline models employing cross-entropy (CE) loss. Oktay \etal~\cite{oktay2018attention} proposed an attention gate to filter the
features propagated through the skip connections of U-Net.

Despite the success of the aforementioned FCN-based methods, they all assume that reliable ground truth annotations are abundant, which is not always the case in practice, not only because collecting pixel-level annotation is time-consuming, but also since human-annotations are inherently noisy. Further, annotations suffer from inter/intra-observer variation even among experts as the boundary of the lesion is often ambiguous. On the other hand, as the high capacity of deep neural networks (DNN) enable them to memorize a random labeling of training data~\cite{zhang2016understanding}, DNNs are potentially exposed to overfitting to noisy labels. Therefore, treating the annotations as completely accurate and reliable may lead to biased models with weak generalization ability. This motivates the need for constructing models that are more robust to label noise.

Previous works on learning a deep classification model from noisy labels can be categorized into two groups. Firstly, various methods were proposed to model the label noise, together with learning a discriminative neural network. For example, probabilistic graphical models were used to discover the relation between data, clean labels and noisy labels, with the clean labels treated as latent variables related to the observed noisy label~\cite{xiao2015learning,vahdat2017toward}. Sukhbaatar~\etal~\cite{sukhbaatar2014learning} and Goldberger~\etal~\cite{goldberger2016training} incorporated an additional layer in the network dedicated to learning the noise distribution. Veit~\etal~\cite{veit2017learning} proposed a multi-task network to learn a mapping from noisy to clean annotations as well as learning a classifier fine-tuned on the clean set and the full dataset with reduced noise. 

Instead of learning the noise model, the second group of methods concentrates on reweighting the loss function. Jiang~\etal~\cite{jiang2017mentornet} utilized a long short-term memory (LSTM) to predict sample weights given a sequence of their cost values. Wang~\etal~\cite{wang2018iterative} designed an iterative learning approach composed of a noisy label detection module and a discriminative feature learning module, combined with a reweighting module on the softmax loss to emphasize the learning from clean labels and reduce the influence of noisy labels. Recently, a more elaborate reweighting method based on a meta-learning algorithm was proposed to assign weights to classification samples based on their gradient direction~\cite{ren2018learning}. A small set of clean data is leveraged in this reweighting strategy to evaluate the noisy samples gradient direction and assign more weights to sample whose gradient is closer to that of the clean dataset.

% comment out
\iffalse
In this work, we aim to extend the idea of example reweighting explored previously for the classification problem to the task of pixel-level segmentation. We propose the first method to learn a set of spatially adaptive weight maps associated with training skin images and adjust the contribution of each pixel in the optimization of deep network. Inspired by Ren~\etal~\cite{ren2018learning}, the importance weights are assigned to pixels based on the pixel-wise loss gradient directions. A meta-learning approach is integrated at every training iteration to approximate the optimal weight maps of the current batch based on the CE loss on a small set of skin lesion images annotated by experts. Learning the deep skin lesion segmentation network and spatially adaptive weight maps are performed in an end-to-end manner. Our experiments show how effective leveraging of a small clean dataset makes a deep segmentation network robust to annotation noise.
\fi

\zm{In this work, we aim to extend the idea of example reweighting~\cite{ren2018learning} explored previously for the classification problem to the task of pixel-level segmentation. We propose the first deep robust network to target the segmentation task by considering the spatial variations in the quality of pixel-level annotations. We learn spatially adaptive weight maps associated with training images and adjust the contribution of each pixel in the optimization of deep network. The importance weights are assigned to pixels based on the pixel-wise loss gradient directions. A meta-learning approach is integrated at every training iteration to approximate the optimal weight maps of the current batch based on the CE loss on a small set of skin lesion images annotated by experts. Learning the deep skin lesion segmentation network and spatially adaptive weight maps are performed in an end-to-end manner. Our experiments show how efficient leveraging of a small clean dataset makes a deep segmentation network robust to annotation noise.}

\section{Methodology}
Our goal is to leverage a combination of a small set of expensive expert-level annotations as well as a large set of unreliable noisy annotations, acquired from, e.g., novice dermatologists or crowdsourcing platforms, into the learning of a fully convolutional segmentation network.

%\cmmnt{we want to find a model theta based on Dc and Dn. We define two loss functions Lc and Ln, where the latter encode level of confidence/noise/weight on samples W. batch bn, update theta based on Ln. batch bc to update W based on Lc.}

\bigbreak
\noindent\textbf{FCN's average loss.}~~In the setting of supervised learning, with the assumption of the availability of high-quality clean annotations for a large dataset of $N$ images and their corresponding pixel-wise segmentation maps, $\mathcal{D}:\{(X(i), Y(i));\\ i = 1,2, \dots , N\}$, parameters $\theta$ of a fully convolutional segmentation network are learned by minimizing the negative log-likelihood of the generated segmentation probability maps in the cost function $\mathcal{L}$: 
\vspace{-0.5em}
\begin{eqnarray}
\mathcal{L}(X,Y;\theta)=-\frac{1}{N} \Sigma_{i=1}^{N}\frac{1}{P}\Sigma_{p\in \Omega_i} y_{p} \log Pr(y_{p}|x_{p};\theta)
\label{orig_loss}
\end{eqnarray}
\noindent where $P$ is the number of pixels in an image, $\Omega_i$ is the pixel space of image $i$, $x_p$ and $y_p$ refer, in order, to the image pixel $p$ and its ground truth label, and $Pr$ is the predicted probability. As the same level of trust in the pixel-level annotations of this clean training data annotations is assumed, the final value of the loss function is averaged equally over all pixels of the training images.

\bigbreak
\noindent\textbf{FCN's weighted loss.}~~As opposed to the fully supervised setting, when the presence of noise in most training data annotations is inevitable while only a limited amount of data can be verified by human experts, our training data comprises of two sets: $\mathcal{D}^c:\{(X^c(i), Y^c(i)); i = 1,2, \dots , K\}$ with verified clean labels and $\mathcal{D}^n:\{(X^n(i), Y^n(i)); i = 1,2, \dots , M\gg K\}$ with unverified noisy labels. We also assume that $\mathcal{D}^c \subset \mathcal{D}^n$.
Correspondingly, we have two losses, $\mathcal{L}^c$ and $\mathcal{L}^n$. Whereas $\mathcal{L}^c$ has equal weighting, $\mathcal{L}^n$ penalizes a log-likelihood of the predicted pixel probabilities but \emph{weighted} based on the amount of noise:
\vspace{-0.5em}
\begin{eqnarray}
{\mathcal{L}}^c(X^c(i),Y^c(i);\theta)=-\frac{1}{P}\Sigma_{p\in \Omega_i} y_{p}^c \log Pr(y_{p}^{c}|x_{p}^{c};\theta),~~~~~\\
%\end{eqnarray}
%\begin{eqnarray}
\mathcal{L}^n({X^n(i),Y^n(i);\theta, W(i)})=-\Sigma_{p\in \Omega_i} y_{p}^n w_{ip} \log Pr(y_{p}^n|x_p^n;\theta)
\label{eqn:Ln}
\end{eqnarray}

\noindent where $w_{ip}$ is the weight associated with pixel $p$ of image $i$. All the weights of the $P$ pixels of image $i$ are collected in a spatially adaptive weight map $W(i)=\{w_{i1},\dots,w_{ip},\dots,w_{iP}\}$, and weight maps associated with all $M$ noisy training images $X^n$ are collected in $W=\{W(1),\dots, W(M)\}$.
\cmmnt{
\begin{figure*}[!t]
\vspace{-1em}
\centering
\includegraphics[width=4.8in,trim={0 3.2in 0 3.2in},clip]{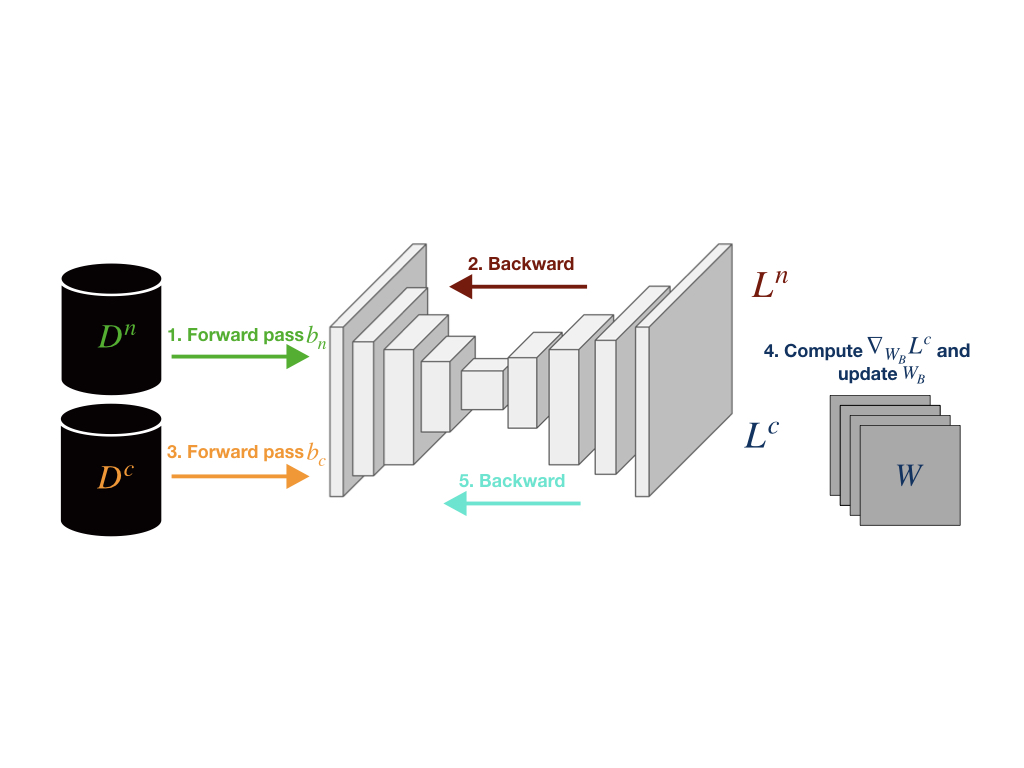}
\vspace{-1em}
\caption{\zm{Learning iterations of the deep segmentation network and spatially adaptive weight maps consists of five steps. A batch of all images is passed through the network with known parameters $\theta_t$~(step 1), $\mathcal{L}^n$ is computed and network parameters are updated to $\hat{\theta}$~(step 2). Next, a batch of clean data is passed through the network with known parameters $\hat{\theta}$~(step 3), $\mathcal{L}^c$ and $\nabla_{W_B}\mathcal{L}^c$ are computed and $W_B$ gets updated~(step 4). Finally, $\mathcal{L}^n$ is computed and network parameters are updated to $\theta_{t+1}$~(step 5).}}
\label{diagram}
\vspace{-1em}
\end{figure*}
}

\noindent\textbf{Model optimization.}~~The deep noise-robust network parameters $\theta$ are now found by optimizing the weighted objective function $\mathcal{L}^n$ (as opposed to equal weighting in~(\ref{orig_loss})) on the noisy annotated data $\mathcal{D}^n$, as follows:
\begin{eqnarray}
\theta^\ast=\argmin_\theta \Sigma_{i=1}^{M}\mathcal{L}^n({X^n(i),Y^n(i);\theta, W(i)}).
\label{optimal_model}
\end{eqnarray} 

\noindent\textbf{Optimal spatially adaptive weights.}~~ The optimal value of unknown parameters $W$ is achieved by minimizing the expectation of negative log-likelihoods in the meta-objective function $\mathcal{L}^c$ over the clean training data $\mathcal{D}^c$:
\begin{eqnarray}
W^\ast= \argmin_{W,~W\geqslant0}\dfrac{1}{K}\Sigma_{i=1}^K {\mathcal{L}}^c(X^c(i),Y^c(i);\theta^*(W)).
\label{optimal_weights}
\end{eqnarray}

%\cmmnt{In this scenario, to learn the network parameters $\theta$, we define and minimize a new loss $\mathcal{L}^n$, which encodes a \emph{weighted} log-likelihood of probabilities over the noisy data set $\mathcal{D}_n$, and leverage the clean data $\mathcal{D}^c$ to determine the weights by minimizing $\mathcal{L}^c$ on the clean data.}

\noindent\textbf{Efficient meta-training.}~~Solving (\ref{optimal_weights}) to optimize the spatially adaptive weight maps $W$ for each update step of the network parameter $\theta$ in~(\ref{optimal_model}) is inefficient. Instead, an online meta-learning approach is utilized to approximate $W$ for every gradient descent step involved in optimizing $\theta$ (\ref{optimal_model}). At every update step $t$ of $\theta$ (\ref{optimal_model}), we pass a mini-batch $b_n$ of noisy data forward through the network and then compute one gradient descent step toward the minimization of $\mathcal{L}^n$:

\begin{eqnarray}
\vspace{-1em}
\hat{\theta}=\theta_t - \alpha \nabla_{\theta} \Sigma_{i=1}^{|b_n|} \mathcal{L}^n(X^n(i),Y^n(i);\theta_t,W_0(i))
\end{eqnarray}

\noindent where $\alpha$ is the gradient descent learning rate and $W_0$ in the initial spatial weight maps set to zero. Next, a mini-batch $b_c$ of clean data is fed forwarded through the network with parameters $\hat{\theta}$ and the gradient of ${\mathcal{L}}^c$ with respect to the current batch weight maps $W^B=\{W(1),\dots,W(|b_n|)\}$ is computed. We then take a single step toward the minimization of ${\mathcal{L}}^c$, as per (\ref{optimal_weights}), and pass the output to a rectifier function as follows:

\begin{eqnarray}
U^B= W_0^B \bigg\rvert_{W_0^B=\textbf{0}}-\eta\nabla_{W^B}\dfrac{1}{|b_c|}\Sigma_{i=1}^{|b_c|} {\mathcal{L}}^c(X^c(i),Y^c(i);\hat{\theta}(W)),\label{weight_gradients}\\
%\end{eqnarray}
%\vspace{-1em}
%\begin{eqnarray}
W^B= g(\max(\textbf{0},U^B)).~~~~~~~~~~~~~~~~~~~~~~~~~
\label{weight_rectify}
\end{eqnarray}

\noindent \zm{where $\eta$ is a gradient descent learning rate, $\max$ is an element-wise max and $g$ is the normalization function. Following the average loss over a mini-batch samples in training a deep network, $g$ normalizes the learned weight maps such that $\Sigma_{i=1}^{|b_n|}\Sigma_{p\in\Omega_i}w_{ip}=1$.}

Equations (\ref{weight_gradients}) and (\ref{weight_rectify}) clarify how the learned weight maps prevents penalizing the pixels whose gradient direction is not similar to the direction of gradient on the clean data. A negative element $u_{ip}$ in $U$ (associated with pixel $p$ of image $i$) implies a positive gradient $\nabla_{w_{ip}}\mathcal{L}^c$ in (\ref{weight_gradients}), meaning that increasing the assigned weight to pixel $p$, $w_{ip}$, increases the $\mathcal{L}^c$ loss value on clean data. So by rectifying the values of $u_{ip}$ in~(\ref{weight_rectify}), we assign zero weights $w_{ip}$ to pixel $p$ and prevent penalizing it in the loss function. In addition, the rectify function makes the $\mathcal{L}^n$ loss non-negative (cf.~(\ref{eqn:Ln})) and results in more stable optimization.

Once the learning of spatially adaptive weight maps is performed, a final backward pass is needed to minimize the reweighted objective function and update the network parameters from $\theta_t$ to $\theta_{t+1}$:
%\vspace{-0.5em}
\begin{eqnarray}
\theta_{t+1}=\theta_t - \alpha \nabla_{\theta_t} \Sigma_{i=1}^{|b_n|} \mathcal{L}^n(X^n(i);\theta_t,W^B).
\end{eqnarray}
\cmmnt{\fig{\ref{diagram}} shows the learning overview of the network parameters $\theta$ as well as spatial weight maps $W$.}

\section{Experiments and Discussion}
\noindent\textbf{Data description.}~~We validated our spatially adaptive reweighting approach on data provided by the International Skin Imaging Collaboration (ISIC) in 2017~\cite{codella2017skin}. The data consists of $2000$ training, $150$ validation and $600$ test images with their corresponding segmentation masks. The same split of validation and test data are deployed for setting the hyper-parameters and reporting the final results. We re-sized all images to $96\times96$ pixels and normalized each RGB channel with the per channel mean and standard deviation of training data.

\begin{figure*}[!h]
\vspace{-2em}
\centering
\includegraphics[width=4.8in,trim={0 0.3in 0 0.3in},clip]{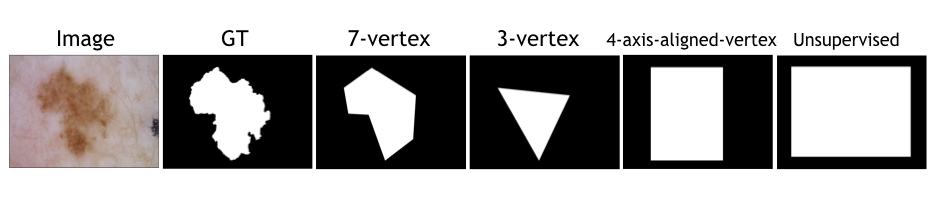}
\vspace{-1em}
\caption{A skin image and its clean and various noisy segmentation maps.}
\label{polygons}
\vspace{-3.5em}
\end{figure*}

To create noisy ground truth annotations, we consider a lesion boundary as a closed polygon and simplify it by reducing its number of vertices:  Less important vertices are discarded first, where the importance of each vertex is proportional to the acuteness of the angle formed by the two adjacent polygon line segments and their length. $7$-vertex, $3$-vertex and $4$-axis-aligned-vertex polygons are generated to represent different levels of annotation noise for our experiments.  To simulate an unsupervised setting, as an extreme level of noise, we automatically generated segmentation maps that cover the whole image (excluding a thin band around the image perimeter). \fig{\ref{polygons}} shows a sample lesion image and its associated ground truth as well as generated noisy annotations.

\bigbreak
%\begin{sloppypar}
\noindent\textbf{Implementation.}~~We utilize PyTorch framework to implement our segmentation reweighting network. We adopt the architecture of fully convolutional network U-Net~\cite{ronneberger2015u} initialized by a random Gaussian distribution. We use the stochastic gradient descent algorithm for learning the network parameters from scratch as well as the spatial weight maps over the mini-batch of sizes $|b_n|=2$ and $|b_c|=10$. We set the initial learning rate for both $\alpha$ and $\eta$ to $10^{-4}$ and divide by $10$ when the validation performance stops improving. We set the momentum and weight decay to $0.99$ and $5\times10^{−5}$, respectively. Training the deep reweighting network took three days on our $12$ GB GPU memory.\\
%\end{sloppypar}

\begin{figure*}[!t]
\vspace{-1em}
\centering
%\captionsetup{skip=0pt, font=small}
\includegraphics[width=4.8in]{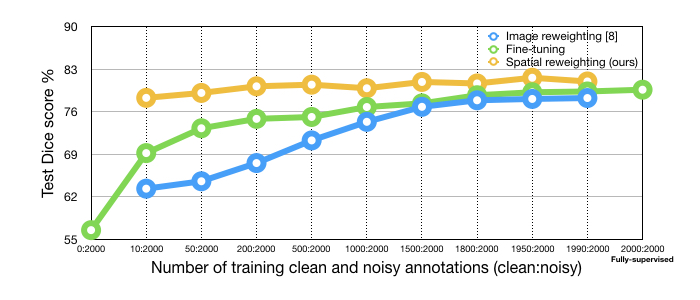}
\vspace{-2em}
\caption{\zm{Test Dice score comparison for fine-tuning, per image reweighting~\cite{ren2018learning} and, spatially adaptive reweighting (ours) models.}}
\label{ft_vs_rw}
\vspace{-2em}
\end{figure*}

\noindent\textbf{Spatially adaptive reweighting vs. image reweighting and fine-tuning.} We compare our work with previous work on noisy labels which assign a weight per training images~\cite{ren2018learning}. In addition, one popular way of training a deep network when a small set of clean data as well as a large set of noisy data are available is to pre-train the network on the noisy dataset and then fine-tune it using the clean dataset. By learning the spatially adaptive weight maps proposed in this work, we expect to leverage clean annotations more effectively for segmentation task and achieve an improved performance. We start with $|\mathcal{D}^n|=2000$ images annotated by $3$-vertex polygons and gradually replace some of the noisy annotation with expert-level clean annotations, i.e., increase $|\mathcal{D}^c|$. We report the Dice score on the test set in~\fig{\ref{ft_vs_rw}}. The first (leftmost) point on the fine-tuning curve indicates the result of U-Net when all annotation are noisy and the last point corresponds to a fully-supervised U-Net. When all annotation are either clean or noisy, training the reweighting networks are not applicable. We observe a consistent improvement in the test Dice score when the proposed reweighting algorithm is deployed. In particular, a bigger boost in improvement when the size of the clean annotation is smaller signifies our method's ability to effectively utilize even a handful of clean samples.\\

\noindent\textbf{Size of the clean dataset.}~~\fig{\ref{ft_vs_rw}} shows the effect of the clean data size, $|\mathcal{D}^c|$, on the spatial reweighting network performance. Our results show leveraging just $10$ clean annotations in the proposed model improves the test Dice score by $21.79\%$ in comparison to training U-Net on all noisy annotations. Also, utilizing $50$ clean annotations in the spatial reweighting algorithm achieves a test Dice score ($\sim$80\%) almost equal to that of the fully supervised approach. With only $\sim$100 clean image annotations, the spatial reweighting method outperforms the fully-supervised with 2000 clean annotations. Incrementing $|\mathcal{D}^c|$ from 50 to 1990, the reweighting approach improves the test Dice score by about $2\%$, questioning whether a $2\%$ increase in accuracy is worth the $\sim$40-fold increase in annotation effort. Outperforming the supervised setting using spatial reweighting algorithm suggests that the adaptive loss reweighting strategy works like a regularizer and improves the generalization ability of the deep network.

\begin{table*}[!t]
\renewcommand{\arraystretch}{1}
\centering
%\vspace{-2em}
\caption{Dice score using fine-tuning and reweighting methods for various noise levels.}
%\scriptsize
\vspace{-1em}
\begin{center}
%\resizebox{0.9\textwidth}{!}{
\begin{tabular}{|c|l|c|c|c|}
\hline
\textbf  &noise type & fine-tuning & proposed reweighting \\
\hline
~A~ & no noise (fully-supervised)& 78.63\% & not applicable \\
\hline
~B~ & $7$-vertex& 76.12\% & 80.72\% \\
\hline
~C~ & $4$-axis-aligned-vertex & 75.04\% & 80.29\%  \\
\hline
~D~ & $3$-vertex & 73.02\% &  79.45\% \\
\hline
~E~ & maximal (unsupervised) & 70.45\% & 73.55\%  \\
\hline
\end{tabular}
%}
\end{center}
\label{noise_robustness}
\vspace{-2em}
\end{table*}

\begin{figure*}[!h]
\centering
%\vspace{-1em}
\captionsetup{skip=3pt, font=small}
\includegraphics[width=4.9in]{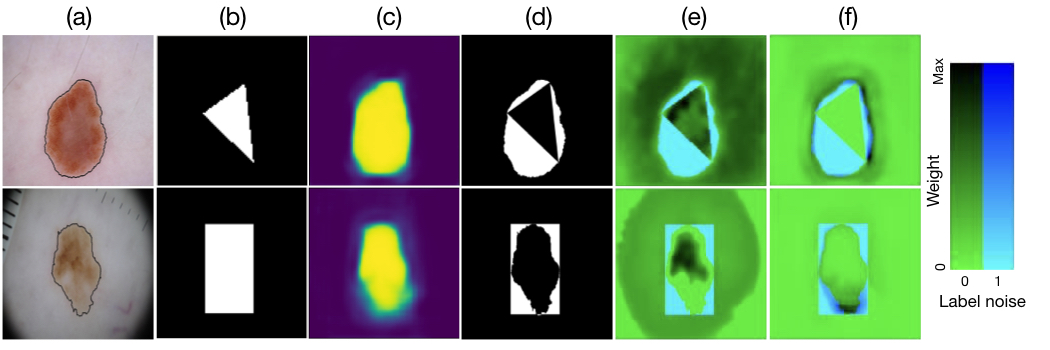}
\caption{(a) Sample skin images and expert lesion delineations (thin black contour), (b) noisy ground truth, (c) network output, (d) the erroneously labelled pixels (i.e. noisy pixels) and learned weight maps in iterations (e) 1K and (f) 100K overlaid over the noisy pixel masks using the following coloring scheme: Noisy pixels are rendered via the blue channel: mislabelled pixels are blue, and weights via the green channel: the \emph{lower} the weight the greener the rendering. The cyan color is produced when mixing green and blue, i.e. when low weights (green) are assigned to mislabelled pixels (blue). Note how the cyan very closely matches (d), i.e. mislabelled pixels are ca. null-weighted.}
\label{weight_maps}
\vspace{-2em}
\end{figure*}

\bigbreak
\noindent\textbf{Robustness to noise.}~~In our next experiment, we examine how the level of noise in the training data affect the performance of the spatial reweighting network in comparison to fine-tuning. We utilized four sets of (i) $7$-vertex; (ii) $3$-vertex; (iii) $4$-axis-aligned-vertex simplified polygons as segmentation maps; and (iv) unsupervised coarse segmentation masks where each set corresponds to a level of annotation noise (Fig.~\ref{polygons}). Setting $|\mathcal{D}^c|=100$ and $|\mathcal{D}^n|=1600$, the segmentation Dice score of test images for reweighting and fine-tuning approaches are reported in~\tab{\ref{noise_robustness}}. We observe that deploying the proposed reweighting algorithm for 3-vertex annotations outperforms learning from accurate delineation without reweighting. Also, increasing the level of noise, from 7-vertex to 3-vertex polygon masks in noisy data, results in just $\sim$1\% Dice score drop when deploying reweighting compared to $\sim$3\% drop in fine-tuning.\\
\cmmnt{***drop is ~2.5\% from accurate delineation to 7 vertex***}

\noindent\textbf{Qualitative results.}~~To examine the spatially adaptive weights more closely, for some sample images, we overlay the learned weight maps, in training iterations 1K and 100K, over the incorrectly annotated pixels mask~(\fig{\ref{weight_maps}}). To avoid overfitting to annotation noise, we expect the meta-learning step to assign zero weights to noisy pixels (the white pixels in~\fig{\ref{weight_maps}}-(d)). Looking into ~\fig{\ref{weight_maps}}-(e,f) confirms that the model consistently learns to assign zero (or very close to zero) weights to noisy annotated pixels (cyan pixels), which ultimatly results in the prediction of the segmentation maps in~\fig{\ref{weight_maps}}-(c) that, qualitatively, closely resemble the unseen expert delineated contours shown in~\fig{\ref{weight_maps}}-(a).\\

\vspace{-2em}
\section{Conclusion}
By learning a spatially-adaptive map to perform pixel-wise weighting of a segmentation loss, we were able to effectively leverage a limited amount of cleanly annotated data in training a deep segmentation network that is robust to annotation noise. We demonstrated, on a skin lesion image dataset, that our method can greatly reduce the requirement for careful labelling of images without sacrificing segmentation accuracy. Our reweighting segmentation network is trained end-to-end, can be combined with any segmentation network architecture, and does not require any additional hyper-parameter tuning.

\vspace{00.8em}
\noindent\textbf{Acknowledgments}. We thank NVIDIA Corporation for the donation of Titan X GPUs used in this research and Anonymous for partial funding.
\vspace{-0.5em}
\bibliographystyle{splncs03}
\bibliography{references}

\end{document}